\PassOptionsToPackage{dvipsnames, svgnames, x11names, table}{xcolor}
\documentclass[table]{article} % For LaTeX2e
\usepackage{iclr2025_conference,times}

\usepackage{xspace}
\usepackage{microtype}
\usepackage[pagebackref=true,breaklinks=true,colorlinks,bookmarks=false]{hyperref}
\usepackage{url}
\usepackage{graphicx}
\usepackage{booktabs}
\usepackage{array}
\usepackage{arydshln}
\usepackage[ruled]{algorithm2e}
\usepackage{multirow}
\usepackage{wrapfig}
\usepackage{float}
\usepackage{svg}
\usepackage{subfig}
\usepackage{adjustbox}
\usepackage{bbding}
\usepackage{lipsum}  
\usepackage{colortbl}
\usepackage{tcolorbox}

\definecolor{color3}{HTML}{e2ecda}

\newcommand{\ie}{{\emph{i.e.}},\xspace}
\newcommand{\eg}{\emph{e.g.},\xspace}
\newcommand{\etc}{etc\@ifnextchar.{}{.\@}}

\title{Dog-IQA: Standard-guided Zero-shot MLLM \\ for Mix-grained Image Quality Assessment}

\author{
  Kai Liu$^{1}$\thanks{Equal contribution}~,\enspace
  Ziqing Zhang$^{1}$\footnotemark[1]~,\enspace
  Wenbo Li$^{2}$,\enspace
  Renjing Pei$^{3}$,\enspace
  Fenglong Song$^{3}$,\enspace
  Xiaohong Liu$^{1}$,\enspace \\
  ~\textbf{Linghe Kong}$^{1}$\footnotemark[2]~,\enspace
  \textbf{Yulun Zhang}$^{1}$\thanks{Corresponding authors: Yulun Zhang, yulun100@gmail.com, Linghe Kong,  linghe.kong@sjtu.edu.cn}\\
  \textsuperscript{1}Shanghai Jiao Tong University,\enspace
  \textsuperscript{2}The Chinese University of Hong Kong,\enspace \\
  \textsuperscript{3}Huawei Technologies Ltd.\enspace 
  \vspace{-8mm}
}

\usepackage{amsmath,amsfonts,bm}

\def\eqref#1{equation~\ref{#1}}
\def\1{\bm{1}}

\def\va{{\bm{a}}}

\def\vs{{\bm{s}}}

\DeclareMathAlphabet{\mathsfit}{\encodingdefault}{\sfdefault}{m}{sl}
\SetMathAlphabet{\mathsfit}{bold}{\encodingdefault}{\sfdefault}{bx}{n}

\iclrfinalcopy % Uncomment for camera-ready version, but NOT for submission.
\begin{document}
\pdfoutput=1
% \vspace{-3mm}
\maketitle
\vspace{-4mm}
\begin{abstract}
\vspace{-3mm}
Image quality assessment (IQA) serves as the golden standard for all models' performance in nearly all computer vision fields.
However, it still suffers from poor out-of-distribution generalization ability and expensive training costs.
To address these problems, we propose \textbf{Dog-IQA}, a stan\textbf{D}ard-guided zer\textbf{o}-shot mix-\textbf{g}rained IQA method, which is training-free and utilizes the exceptional prior knowledge of multimodal large language models (MLLMs).
To obtain accurate IQA scores, namely scores consistent with humans, we design an MLLM-based inference pipeline that imitates human experts.
In detail, Dog-IQA applies two techniques.
\textbf{First}, Dog-IQA objectively scores with specific standards that utilize MLLM's behavior pattern and minimize the influence of subjective factors.
\textbf{Second}, Dog-IQA comprehensively takes local semantic objects and the whole image as input and aggregates their scores, leveraging local and global information.
Our proposed Dog-IQA achieves state-of-the-art (SOTA) performance compared with training-free methods, and competitive performance compared with training-based methods in cross-dataset scenarios.
Our code will be available at \href{https://github.com/Kai-Liu001/Dog-IQA}{https://github.com/Kai-Liu001/Dog-IQA}.
\end{abstract}

% \vspace{-2mm}
\vspace{-6mm}
\section{Introduction}
\vspace{-3mm}
\begin{wrapfigure}{r}{0.45\textwidth}
\vspace{-7mm}
\hspace{-3mm}
\centering
\includegraphics[width=\linewidth]{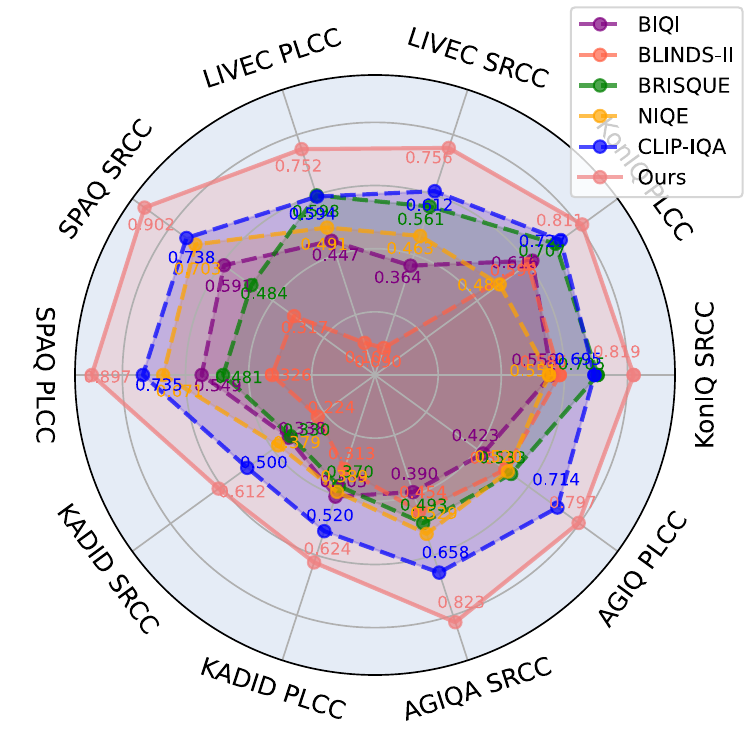}
\vspace{-5mm}
\caption{Comparison between Dog-IQA and existing training-free IQA SOTAs, exhibiting Dog-IQA's excellent zero-shot IQA ability.}
\label{fig:radar-zeroshot}
\vspace{-6mm}
\end{wrapfigure}

Image quality assessment (IQA) aims to provide accurate quality scores that align with human mean opinion scores (MOS).
With the booming of digital technology, the explosion of visual content calls for advanced IQA methods in all fields including communication~\citep{zhou2022quality}, entertainment~\citep{wu2024assessor360}, professional use~\citep{chow2016review,fang2020perceptual}, and recently popular AI-generated content~\citep{pickapic,li2023agiqa}.
Over time, significant contributions have been made in this domain, evolving from traditional handcrafted feature-based approaches~\citep{wang2004image,mittal2012making} to deep neural network (DNN)-based methods~\citep{talebi2018nima,ying2020patches,qin2023data,saha2023re}, bringing steady improvements in accuracy.

Nonetheless, these IQA methods still suffer from poor out-of-distribution (OOD) generalization ability~\citep{you2024descriptive} and expensive training costs~\citep{wu2024q}.
One potential solution to the OOD issue involves training DNNs on a combination of multiple IQA datasets. 
Although it sounds promising, this approach fails due to inconsistent standards used during dataset construction, leading to distribution mismatches across datasets. 
For instance, an image rated high quality in one dataset may receive a low-quality score in another, ultimately degrading model performance. 
Another approach is to create a larger, more diverse dataset representing a wide range of distortions. 
However, aside from the increased training costs, the scoring process is labor-intensive and time-consuming, making this approach impractical. 
As a result, poor OOD performance remains an open problem.

Recently, MLLMs have shown impressive zero-shot capabilities across various computer vision tasks, including classification~\citep{radford2021learning}, segmentation~\citep{li2024groundinggpt,he2024weakly}, detection~\citep{zhang2023superyolo}, and restoration~\citep{chen2023image,zhao2024equivariant}.
Thanks to their extensive training on large datasets and vast model sizes~\citep{liu2024visual,awadalla2023openflamingo}, MLLMs possess rich prior knowledge and are closely aligned with human perceptual understanding~\citep{yin2023survey}.
As the MLLM has not been trained on IQA-related datasets, previous related research~\citep{wu2024q,wu2024towards} mainly focuses on training or fine-tuning.
These studies have demonstrated remarkable accuracy, suggesting that MLLMs hold great potential for driving the next wave of IQA advancements (see Figure~\ref{fig:radar-zeroshot}).
However, while fine-tuning significantly enhances accuracy, it introduces additional computational costs and complexity. 
Therefore, we aim to fully exploit MLLMs' potential \textbf{without resorting to fine-tuning or task-specific training}.

\begin{figure}[t]
    \vspace{-4mm}
    \centering
    \includegraphics[width=0.95\linewidth]{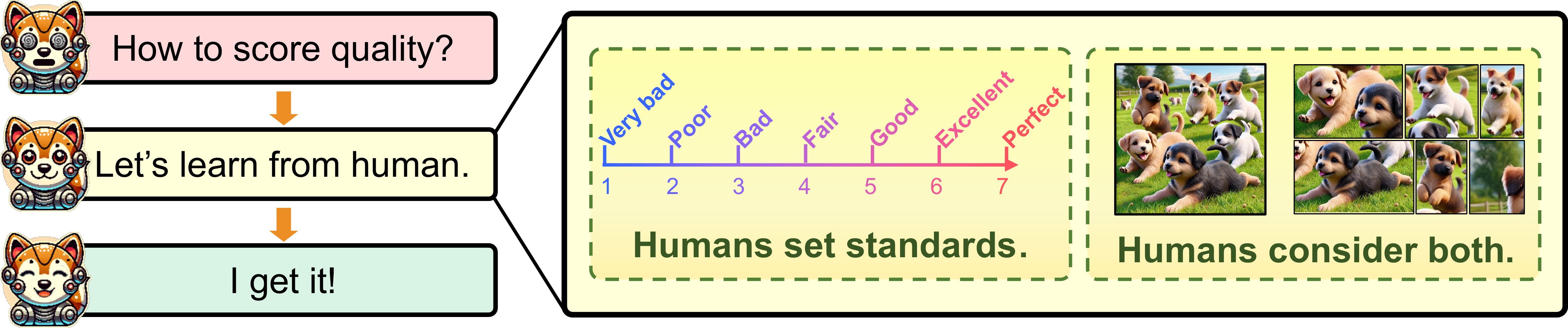}
    \vspace{-2mm}
    \caption{
    The idea of Dog-IQA is inspired by the human evaluator's scoring procedures. 
    When scoring, human evaluators are provided with standards mapping the quality to scores.
    Then they start with the global quality and zoom in on objects to grasp local quality. 
    We integrate these key procedures and switch their form according to MLLM's behavior pattern, formulating Dog-IQA.
    }
    \label{fig:human}
    \vspace{-5mm}
\end{figure}

Our approach is inspired by the human evaluators' scoring process and the MLLMs' behavior pattern~\citep{yin2023survey}. 
Thus, we design an inference pipeline mimicking human image scoring which is shown in Figure~\ref{fig:human}. 
Our key designs stem from the following observations. 
\textbf{First}, when human evaluators score images, they are typically provided with a clear standard for each quality level~\citep{wu2023q_align}. 
Without such a standard, discrepancies arise—for example, one person may interpret a score of 60 as merely passing, while another views 50 as average. 
By providing a consistent scoring standard, evaluators are more likely to agree on quality assessments. 
\textbf{Second}, when humans assess image quality, they consider both global and local quality~\citep{navon1977forest,gerlach2018navon}, often zooming in to evaluate specific areas~\citep{forster2012glomosys}. 
Notably, these zoomed-in evaluations are typically centered on objects within the image rather than being performed randomly.
\textbf{Additionally}, MLLMs generate outputs in token form, making it difficult for them to produce precise scores, such as 86.5, which would require generating multiple tokens. 

Building on these observations, we propose two novel techniques. 
\textbf{First}, we develop a standard-guided scoring system that aims to establish a clear mapping between quality levels and scores and restrict the MLLM to scoring within a predefined range $\{1, 2, \dots, K\}$.
The mapping and restriction ensure the model's understanding of the quality scale. 
\textbf{Second}, we utilize segmentation models to provide MLLM with the whole image and object-centered sub-images. We then aggregate the scores using an area-weighted average approach.
Our key contributions can be summarized as follows:

\begin{itemize}
    \item We propose \textbf{Dog-IQA}, a standard-guided mix-grained IQA framework that does not require any task-specific training or fine-tuning. Dog-IQA fully leverages the inherent capabilities of pre-trained MLLM and segmentation model to provide accurate IQA scores. Our Dog-IQA serves as a new paradigm for training-free approaches in IQA tasks.
    \item 
    We design two key mechanisms to enhance IQA performance. The standard-guided scoring mechanism ensures consistent and objective quality evaluation by aligning scores with predefined standards. The mix-grained aggregation mechanism refines the final quality score by aggregating global and object-centered sub-image quality scores. 
    \item 
    We conduct extensive experiments and compare Dog-IQA against SOTA IQA methods across multiple datasets. 
    The main experiments show that our proposed Dog-IQA achieves SOTA performance compared with training-free methods, and competitive performance compared with training-based methods in cross-dataset scenarios. 
    
\end{itemize}

\vspace{-3mm}
\section{Related Works}
\vspace{-3mm}
\noindent \textbf{Training-free IQA.}
Training-free IQA is a critical approach in the field of image processing, allowing for the evaluation of image quality without the need for distortion-specific or human-rated training data. Traditional training-free IQA methods are often based on the statistical properties of images, focusing on full-reference (FR) metrics such as PSNR and SSIM~\citep{wang2004image}. 
As for no-reference (NR) training-free IQA, NIQE~\citep{mittal2012making} assesses image quality through the analysis of natural scene statistics features and provides robust but less precise results.
In recent years, CLIP~\citep{radford2021learning}, a multimodal model, has emerged as a significant player, providing robust training-free performance support for prevalent deep-learning-based IQA. 
CLIP-IQA~\citep{wang2023exploring} explores the capabilities of CLIP for assessing image quality and aesthetic perception and pioneers the use of contrastive prompt strategies for scoring. 
ZEN-IQA~\citep{miyata2024zen} and GRepQ~\citep{srinath2024learning} also harness CLIP, with ZEN-IQA utilizing antonym prompts and GRepQ combining low-level and high-level feature representations for IQA. 
While these developments represent a substantial leap forward, there is still significant potential for enhancing the performance of training-free IQA models in terms of accuracy and interpretability.

\noindent \textbf{MLLMs for IQA.}
High-performance MLLMs, such as mPLUG-Owl~\citep{ye2023mplug,ye2024mplug_2,ye2024mplug_3}, LLaVA~\citep{liu2024visual,liu2024improved,liu2024llava}, and InternLM-XComposer~\citep{zhang2023internlm,dong2024internlm}, can be exceptionally utilized to align IQA tasks with human perception.
Based on a comprehensive study~\citep{wu2024comprehensive}, recent efforts concentrate on benchmarking and fine-tuning MLLMs for IQA.
Q-Bench~\citep{wu2023q_bench} and DepictQA~\citep{you2024descriptive} establish evaluation benchmarks for the perceptual, descriptive, comparative, and evaluative capabilities of MLLMs in low-level vision. 
Based on these works, Q-Instruct~\citep{wu2024q} and Co-Instruct~\citep{wu2024towards} further advance the low-level perceptual and descriptive capabilities of MLLMs by introducing large-scale datasets and conducting pre-training. 
Q-Align~\citep{wu2023q_align} categorizes image quality into five tiers, enabling more precise quality score regression. 
However, the cost of fine-tuning large models is substantial, prompting the consideration of more efficient approaches.

\vspace{-4mm}
\section{Methodology}
\vspace{-4mm}

\begin{figure}
    \centering
    \includegraphics[width=\linewidth]{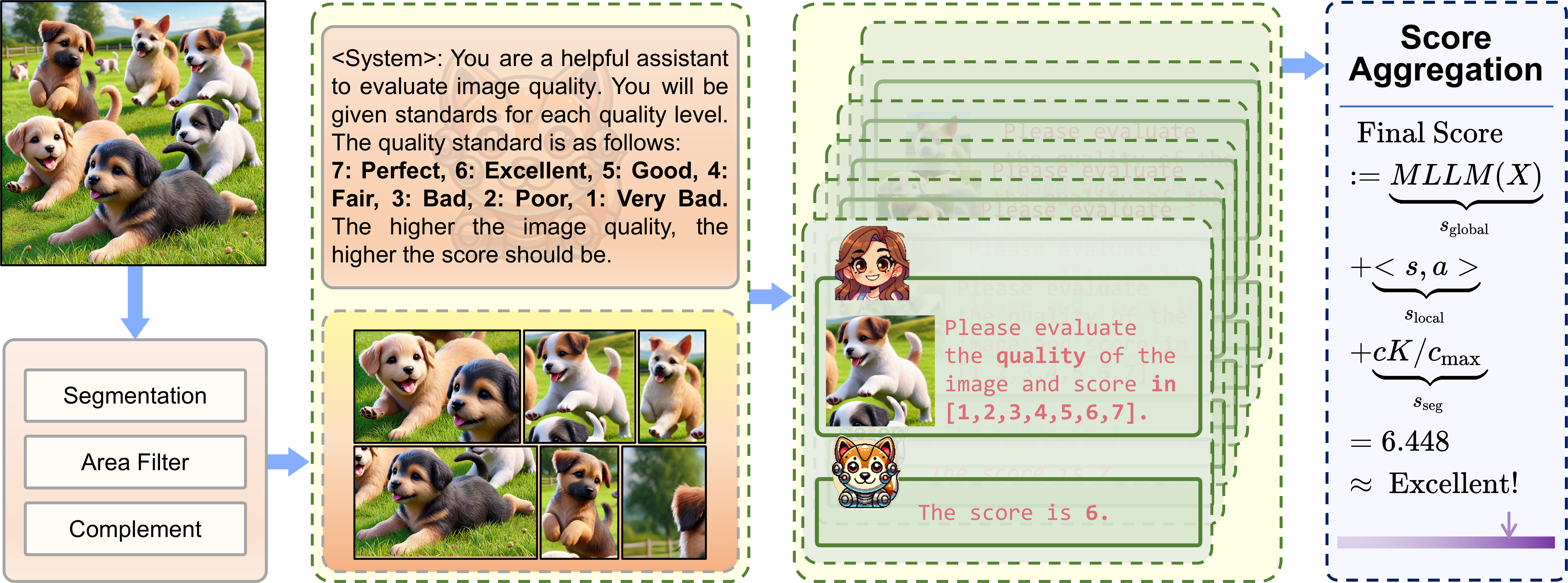}
    \vspace{-5mm}
    \caption{The overall pipeline for our proposed Dog-IQA. It can be divided into three stages, \ie segmentation, standard guided scoring, and score aggregation. The input image is segmented into multiple sub-images centered on objects. Then, MLLM scores with quality standards. After the area-weighted average and addition with $s_{seg}$, the scores are aggregated as the final quality score.}
    \label{fig:pipeline}
    \vspace{-4mm}
\end{figure}

We provide a comprehensive explanation of our proposed Dog-IQA method. 
The overall pipeline of our proposed Dog-IQA is shown in Figure~\ref{fig:pipeline}.
The image to be assessed is segmented into multiple sub-images with the segmentation process pipeline.
Given a detailed standard, the MLLM rates the whole image and sub-images with scores in $\{1,2,\dots,7\}$.
These scores will be finally aggregated to form the final number.
Specifically, we first propose the standard-guided scoring mechanism, which effectively leverages its prior knowledge and its behavior pattern.
Second, we discuss the mix-grained aggregation mechanism, which consists of the process of obtaining suitable sub-images and the aggregation of scores.
The rationale behind using sub-images as inputs is also included.

\vspace{-2mm}
\subsection{Standard-guided Scoring Mechanism}
\vspace{-2mm}
The ultimate goal of image quality assessment (IQA) is to evaluate images in a manner that closely mirrors human judgment.
Thanks to their extensive training data and vast prior knowledge, MLLMs are capable of perceiving images in a way that aligns with human perception~\citep{wu2023q_bench}, giving them an inherent advantage for IQA tasks.
However, expecting an MLLM to output precise quality scores, such as 87.5, is impractical. This is because a score like 87.5 is not represented by a single token, but by four separate tokens: 8, 7, dot, and 5 respectively. Typically, MLLMs can hardly grasp the internal relationship between these tokens, making it difficult for them to associate these values with image quality.
These observations and analyses motivate us to \textbf{insight 1}: 

\textit{It is more effective to represent image quality using one single token to achieve an accurate score.}

Additionally, relying solely on numeric outputs may not be the most optimal approach for two key reasons. 
First, numbers constitute only a small fraction of the data within the training set compared to textual information. 
However, using only text is also not feasible, as we still need to extract a quantitative score. 
Second, human interpretation of numeric scores can vary. 
For instance, some may consider a score of 60 to be just passing, while others may view 50 as an average score. 
Therefore, when human evaluators score image quality, they are often provided with clear standards for each level of quality~\citep{wu2023q_align}.
This observation brings us to \textbf{insight 2}: 

\textit{A combination of text and numbers is a more effective prompt format for MLLM IQA.}

In our proposed method, we integrate these two insights and design the prompt as follows:
\begin{tcolorbox}[colback=yellow!10, colframe=black, boxrule=0.5mm]
\textit{\# System:} {\tt <img>} \textit{You are a helpful assistant to evaluate image quality. You will be given standards for each quality level. The quality standard is listed as follows: \textbf{7: Perfect, 6: Excellent, 5: Good, 4: Fair, 3: Bad, 2: Poor, 1: Very Bad}. The higher the image quality is, the higher the score should be.}

\textit{\# User:} {\tt <img>} \textit{Please evaluate the quality of the image and score in \textbf{[1, 2, 3, 4, 5, 6, 7]}.}
\end{tcolorbox}

In our method, MLLM only outputs discrete numbers from 1 to 7.
While this discrete scoring approach may introduce a slight loss in precision compared to continuous values, the impact is minimal.
Denote that integer score as $s\in \{s|s\in \mathbb{Z}^+ \wedge 1\leq s \leq K \}$, the ground truth MOS as $s^{*}$, and the maximal and minimal value of $s^{*}$ as $\text{Max}_{gt}$ and $\text{Min}_{gt}$ respectively.
We scale $s^{*}$ to $\{1,2,\dots,K\}$ and round it to the nearest integer.
The conversion formula is expressed as:
\begin{equation}
    \hat{s}^{*} = \texttt{Round}((s^{*} - \text{Min}_{gt})/(\text{Max}_{gt} - \text{Min}_{gt})\times (K-1) ) .
\end{equation}
\begin{wraptable}{r}{0.450\textwidth}
\centering
\small
\vspace{-4mm}
\caption{The approximation of performance upper bound of using only K integers to score. The value is calculated by $(SRCC+PLCC)/2$.}
\vspace{-2mm}
% \resizebox{\linewidth}{!}{
\setlength{\tabcolsep}{1.1mm}
\begin{tabular}{c|ccccc}
\toprule
\rowcolor{color3} K & SPAQ  & KonIQ & LIVEC & AGIQA & KADID \\
\midrule
3 & 0.912 & 0.830 & 0.915 & 0.923 & 0.942 \\
5 & 0.968 & 0.946 & 0.964 & 0.973 & 0.980 \\
7 & 0.983 & 0.967 & 0.982 & 0.986 & 0.988 \\
9 & 0.990 & 0.979 & 0.989 & 0.991 & 0.993 \\
\bottomrule
\end{tabular}
% }
\vspace{-5mm}
\label{tab:switch}
\end{wraptable}
As shown in Table~\ref{tab:switch}, the performance upper bounds for different values of $K$ demonstrate that even when using a limited number of discrete levels, the results surpass those of existing methods. The precision loss introduced by the conversion to discrete scores is minimal and can be considered negligible.

In conclusion, for each image $X_i$, MLLM processes its corresponding segmented masks $M_i$ as input.
For each mask $m_k \in M_i$, MLLM will predict a score $s_k$ from the set $\{1,2,\dots,K\}$.
These individual scores are then compiled into a score list $\vs_i$, which is subsequently used to compute the final quality score.

\vspace{-2mm}
\subsection{Mix-grained Aggregation Mechanism}
\vspace{-2mm}
The mix-grained aggregation mechanism can be divided into two parts.
The first part introduces the segmentation pipeline, while the second part presents the aggregation of multiple scores.

\textbf{Segmentation Process Pipeline.}
When humans recognize an image, they start from the global structure and gradually dive into the local parts.~\citep{navon1977forest,forster2012glomosys,gerlach2018navon}
This hierarchical process also applies when assessing image quality. 
Therefore, under the assumption that MLLMs share a similar perception process, it is essential to deliberately leverage meaningful sub-images. 
Specifically, `meaningful' means that these sub-images should not be obtained through random cropping but through instance or semantic segmentation techniques.

The segmentation model is an excellent choice as it tends to segment the semantic objects out.
The object segmented by the segmentation model is padded with zeros around.
While this padding has minimal impact on human perception, as humans can easily recognize the black padding as meaningless and mentally disregard it, this is not the case for MLLMs. 
The visual encoder within the MLLM processes the padding as part of the actual image, leading the model to misinterpret the black regions as the real background. 
This misunderstanding can result in distinct errors, such as the MLLM perceiving low contrast when the foreground is dark or concluding that the background is too dark.
Both cases can negatively affect image quality assessment's accuracy.

\begin{wrapfigure}{r}{0.54\textwidth}
\vspace{-5mm}
\hspace{-1mm}
\begin{minipage}{\linewidth}
\begin{algorithm}[H]
\KwData{Dataset $\mathcal{D}=\{X_i\}_{i=1}^{N}$, area\_threshold $t$, pretrained SAM2 $\mathcal{S}$}

\KwResult{Masks $\mathcal{M}=\{M_i\}_{i=1}^{N}$}
$\mathcal{M} \leftarrow []$\;
\ForEach{image $X_i$ in $D$ }{
    $raw\_masks$ $\leftarrow$ $\mathcal{S}(X_i)$\;
    $final\_masks$ $\leftarrow$ $[]$\;
    \ForEach{mask $m$ in raw\_masks }{
        \If{m.area $\geq t$}{
            $final\_masks.append(mask)$\;
        }
    }
    $remain\_mask$ = $ \bigcap \{\neg final\_masks\}$\;
    \If{remain\_mask.area $\geq t$}{
        $final\_masks.append(remain\_mask)$\;
    }
    
    $\mathcal{M}.append(final\_masks)$\;
}
\Return $\mathcal{M}$\;
\caption{Segmentation Process Pipeline}
\label{alg:seg}
\end{algorithm}
% \vspace{-5pt}
\end{minipage}
\vspace{-8mm}
\end{wrapfigure}
To address the above issue, we adopt an alternative approach by padding the segmented areas with the original pixel values instead of padding with zeros.
Besides, the segmented results of most segmentation models are highly fine-grained, namely the size of each sub-image is too small to identify the object.
Furthermore, small objects tend to have lower image quality due to insufficient pixel density, making it difficult to display sharp details. 
To mitigate this, we apply a coarser granularity and establish a minimum threshold $t$ for mask size. 
A side effect of this coarser granularity is that the masks may only cover a portion of the image.
In some extreme cases, the segmentation model may fail to segment any objects from low-quality images. 
To compensate for this problem, we create a new mask for the uncovered portions of all previous masks.
The detailed process is in Algorithm~\ref{alg:seg}.

\textbf{Assessment Score Aggregation.}
For a given image $X_i$, after obtaining its global score $s_{global}^{(i)}$, segmented masks $M_i$, and their corresponding scores $\vs_i$, we proceed to compute the final predicted score.
A simple approach of averaging $\vs$ to determine the final score for $X_i$ yields suboptimal performance.
This is because some blurred objects, although too small to be perceptible to humans, may be disproportionately penalized by MLLMs. 
Even worse, these blurred objects account for a large proportion of most images, leading to an unfairly low score.

To address this, we propose using a weighted average of the scores, where the area of the corresponding masks determines the weights. 
Mathematically, this can be expressed as $s_{local}^{(i)} = <\vs_i,\va_i>$, where $<\cdot,\cdot>$ is inner product and $\va_i$ is the vector representing the areas of the masks in $M_i$.
This approach aligns more closely with human perception, as the dominant object in an image typically occupies the largest region, and its quality represents the image's quality. 

Furthermore, it is well-established that image quality can significantly impact the performance of neural networks in tasks such as classification, segmentation, and detection. 
In the context of mask segmentation, the number of masks in an image can serve as an indicator of image quality. 
For high-quality, sharp images, the clearer structure enables the model to segment more detailed masks. 
Based on this observation, we introduce a segmentation score as an additional component of the final score.
This segmentation score is defined as $s_{seg}=cK/c_{max}$, where $c$ is the number of masks, and $c_{max}$ is the maximum number of masks observed across the entire dataset. 
The normalization by $K/c_{max}$ ensures that $s_{seg}$ remains in the same range as the MLLM scores. 
Consequently, the final predicted score in our proposed Dog-IQA framework is given by $s_{Dog}^{(i)} = (s_{global}^{(i)} + s_{local}^{(i)})/2 + s_{seg}^{(i)}$.
\vspace{-6.5mm}
\section{Experiments}
\vspace{-3mm}
\subsection{Experimental Settings}
\vspace{-2mm}
\textbf{Data and Evaluation.}
We select the following datasets to evaluate our IQA method:  KonIQ~\citep{hosu2020koniq}, LIVE Challenge~\citep{ghadiyaram2015massive}, SPAQ~\citep{fang2020perceptual}, KADID~\citep{lin2019kadid}, and AGIQA~\citep{li2023agiqa}.
KonIQ and SPAQ are large in-the-wild IQA datasets with more than 10k images.
LIVE Challenge is a smaller in-the-wild dataset with 1.1k images.
KADID-10k is a synthetic dataset, while AGIQA-3k focuses on AI-generated images. 
Together, these datasets provide a comprehensive range of image types and quality variations for evaluation.

As our proposed method is training-free, we compare its performance against two categories of approaches.
The first category is training-free methods, including BIQI~\citep{moorthy2010two}, BLINDS-II~\citep{saad2010dct}, BRISQUE~\citep{mittal2012no}, NIQE~\citep{mittal2012making}, and CLIP-IQA~\citep{wang2023exploring}.
The second category is training-based methods such as NIMA~\citep{talebi2018nima}, DBCNN~\citep{dbcnn}, HyperIQA~\citep{su2020blindly}, MUSIQ~\citep{ke2021musiq}, CLIP-IQA+~\citep{wang2023exploring}, and current SOTA model Q-Align~\citep{wu2023q_align}.

All methods are evaluated in cross-dataset scenarios to demonstrate their zero-shot capabilities. 
Comparing training-free methods with training-based methods may seem unfair due to the latter’s systematic training on quality assessment. We still perform these comparisons to showcase the robustness and competitive zero-shot performance of our approach.
The evaluation metrics used are Spearman's rank correlation coefficient (SRCC) and Pearson's linear correlation coefficient (PLCC).
Both metrics are widely used in IQA to assess the correlation between the model's predictions and human judgments, typically represented by MOS~\citep{telecom2000recommendation}.
Both metrics fall within the range of $[-1, 1]$, and the performance is considered better when they have higher absolute values.

\begin{table}[!t]
\centering
\small
\renewcommand\arraystretch{1.2}
\caption{Performance comparison of Dog-IQA with other \textbf{training-free} IQA models on KonIQ, LIVE Challenge, SPAQ, KADID-10k and AGIQA-3k. \textbf{Bold} font indicates the best performance.}
\vspace{-2mm}
\setlength{\tabcolsep}{1.1mm}
\resizebox{\linewidth}{!}{\begin{tabular}{l|cccccccccc}
\toprule
\rowcolor{color3}  & \multicolumn{2}{c}{KonIQ} &\multicolumn{2}{c}{LIVE Challenge} & \multicolumn{2}{c}{SPAQ} &\multicolumn{2}{c}{KADID-10k} &\multicolumn{2}{c}{AGIQA-3k}  \\ \cline{2-11}
\rowcolor{color3}  \multirow{-2}{*}{Methods} &SRCC $\uparrow$&PLCC $\uparrow$ & SRCC $\uparrow$ &PLCC $\uparrow$ & SRCC $\uparrow$&PLCC $\uparrow$ & SRCC $\uparrow$&PLCC $\uparrow$ & SRCC $\uparrow$ &PLCC $\uparrow$ \\
\hline
BIQI~\citep{moorthy2010two}   & 0.559 & 0.616 & 0.364 & 0.447 & 0.591 & 0.549 & 0.338 & 0.405 & 0.390 & 0.423 \\
BLIINDS-II~\citep{saad2010dct}  & 0.585 & 0.598 & 0.090 & 0.107 & 0.317 & 0.326 & 0.224 & 0.313 &  0.454 & 0.510 \\
BRISQUE~\citep{mittal2012no} &   0.705 & 0.707 & 0.561 & 0.598 & 0.484 & 0.481 & 0.330 & 0.370 &0.493 & 0.533 \\
NIQE~\citep{mittal2012making}   & 0.551 & 0.488 & 0.463 & 0.491 & 0.703 & 0.671 & 0.379 & 0.389 & 0.529 & 0.520\\
CLIP-IQA~\citep{wang2023exploring}   & 0.695 & 0.727 &  0.612 & 0.594 & 0.738 & 0.735 & 0.500 & 0.520 &  0.658 & 0.714 \\
\hline
\textbf{Dog-IQA} (Ours)   & \textbf{0.819} & \textbf{0.811} & \textbf{0.756} & \textbf{0.752} & \textbf{0.902} & \textbf{0.897} & \textbf{0.612} & \textbf{0.624} & \textbf{0.823} & \textbf{0.797}  \\
\bottomrule
\end{tabular}}
\label{tab:comp-no-training}
\vspace{-3mm}
\end{table}

\textbf{Implementation Details.}\label{sec:settings}
We select the pre-trained SAM2~\citep{ravi2024sam} as the segmentation model and mPLUG-Owl3~\citep{ye2024mplug_3} as the MLLM.
The hyperparameters of SAM2 were adjusted to achieve the desired segmentation granularity, with detailed configurations in the supplementary material. 
Using these hyperparameters, the average number of masks generated for the SPAQ dataset is $7.22$.
The maximum number of masks is $71$.
For mPLUG-Owl3, we utilize its default hyperparameters across all test sets.
The number of standard words is $K=7$.
In rare cases (less than 0.1\%), when the MLLM does not output a numeric score but words, we set the score to $1$.
Our code is written with Python and PyTorch~\citep{paszke2019pytorch} and runs on NVIDIA RTX A6000 GPU.

\vspace{-2mm}
\subsection{Comparison with State-of-the-Art Methods}
\vspace{-2mm}
We conduct extensive experiments to evaluate the performance of our proposed Dog-IQA model.
The comparisons with SOTA methods are divided into two categories: training-free methods, shown in Table~\ref{tab:comp-no-training}, and training-based methods, as presented in Table~\ref{tab:comp-training}.

Training-free methods can be broadly categorized into two types. 
The first category includes CLIP-IQA, which leverages the prior knowledge of CLIP and generates scores based on the similarity between text and image embeddings. 
The second category consists of models such as BIQI, BLINDS-II, BRISQUE, and NIQE, which rely on hand-crafted features.
As shown in Table~\ref{tab:comp-no-training}, the traditional hand-crafted features often fail to score accurately due to the complex nature of human opinions on image quality.
CLIP-IQA benefits from its prior knowledge and demonstrates higher accuracy than hand-crafted feature-based methods. Our Dog-IQA model consistently achieves superior performance across all metrics and datasets, significantly outperforming existing training-free methods.

\begin{table}[!t]
\centering
\small
\renewcommand\arraystretch{1.2}
\setlength{\tabcolsep}{1.5mm}
\caption{Performance comparison of our model with \textbf{training-based} IQA models in \textbf{cross-dataset} scenarios. The best and second-best performance is indicated by \textbf{bold} and \underline{underlined} respectively.}
\vspace{-2mm}
\resizebox{\linewidth}{!}{\begin{tabular}{l|c|cccccccccc}
\toprule
\rowcolor{color3} Training Set:  KonIQ&$\to$Testing Set:& \multicolumn{2}{c}{SPAQ}  &\multicolumn{2}{c}{AGIQA-3k} &\multicolumn{2}{c}{KADID-10k}  \\ \hline
\rowcolor{color3} Method  &Training-free?&SRCC $\uparrow$&PLCC $\uparrow$ & SRCC $\uparrow$&PLCC $\uparrow$ & SRCC $\uparrow$&PLCC $\uparrow$  \\
\hline
NIMA~\citep{talebi2018nima}             & $\times$ & 0.856 & 0.838 & 0.654 & 0.715 & 0.535 & 0.532  \\
DBCNN~\citep{dbcnn}                     & $\times$ & 0.806 & 0.812 & 0.641 & 0.730 & 0.484 & 0.497  \\
HyperIQA~\citep{su2020blindly}          & $\times$ & 0.788 & 0.791 & 0.640 & 0.702 & 0.468 & 0.506  \\
MUSIQ~\citep{ke2021musiq}               & $\times$ & 0.863 & 0.868 & 0.630 & 0.722 & 0.556 & 0.575 \\
CLIP-IQA+~\citep{wang2023exploring}     & $\times$ & 0.864 & 0.866 & 0.685 & 0.736 & \underline{0.654} & \underline{0.653}  \\
Q-Align~\citep{wu2023q_align}           & $\times$ & \underline{0.887} & \underline{0.886} & \underline{0.735} & \underline{0.772} & \textbf{0.684} & \textbf{0.671} \\
\hline

\textbf{Dog-IQA} (Ours)                 & $\checkmark$ & \textbf{0.902} & \textbf{0.897} & \textbf{0.823} & \textbf{0.797} & 0.612 & 0.624   \\
\bottomrule
\end{tabular}}
\resizebox{\linewidth}{!}{\begin{tabular}{l|c|cccccccccc}
\toprule
\rowcolor{color3} Training Set:  SPAQ&$\to$Testing Set:& \multicolumn{2}{c}{KonIQ}  &\multicolumn{2}{c}{AGIQA-3k} &\multicolumn{2}{c}{KADID-10k}  \\ \hline
\rowcolor{color3} Method  &Training-free?&SRCC $\uparrow$&PLCC $\uparrow$ & SRCC $\uparrow$&PLCC $\uparrow$ & SRCC $\uparrow$&PLCC $\uparrow$  \\
\hline
NIMA~\citep{talebi2018nima}             & $\times$ & 0.733 & 0.788 & 0.534 & 0.630 & 0.399 & 0.480  \\
DBCNN~\citep{dbcnn}                     & $\times$ & 0.731 & 0.758 & 0.459 & 0.518 & 0.490 & 0.508  \\
HyperIQA~\citep{su2020blindly}          & $\times$ & 0.714 & 0.742 & 0.570 & 0.649 & 0.381 & 0.448  \\
MUSIQ~\citep{ke2021musiq}               & $\times$ & 0.753 & 0.680 & 0.564 & 0.675 & 0.349 & 0.429 \\
CLIP-IQA+~\citep{wang2023exploring}     & $\times$ & 0.753 & 0.777 & 0.577 & 0.614 & \underline{0.633} & \underline{0.638}  \\
Q-Align~\citep{wu2023q_align}           & $\times$ & \textbf{0.848} & \textbf{0.879} & \underline{0.723} & \underline{0.786} & \textbf{0.743} & \textbf{0.740} \\
\hline

\textbf{Dog-IQA} (Ours)                 & $\checkmark$ & \underline{0.819} & \underline{0.811} & \textbf{0.823} & \textbf{0.797} & 0.612 & 0.624   \\
\bottomrule
\end{tabular}}
\label{tab:comp-training}
\vspace{-2mm}
\end{table}

Table~\ref{tab:comp-training} summarizes the performance of various training-based methods in cross-dataset evaluations.
These experiments test the out-of-distribution generalization ability of the models, which is a crucial aspect of IQA. 
For these comparisons, we select KonIQ and SPAQ as training sets due to their large size and in-the-wild characteristics. 
Notably, our Dog-IQA method requires \textbf{no training or fine-tuning} on these datasets, making its strong performance even more remarkable.

\begin{figure}[t]
    \vspace{-2mm}
    \centering
    \begin{minipage}{0.48\textwidth}
        \centering
        \includegraphics[width=\linewidth]{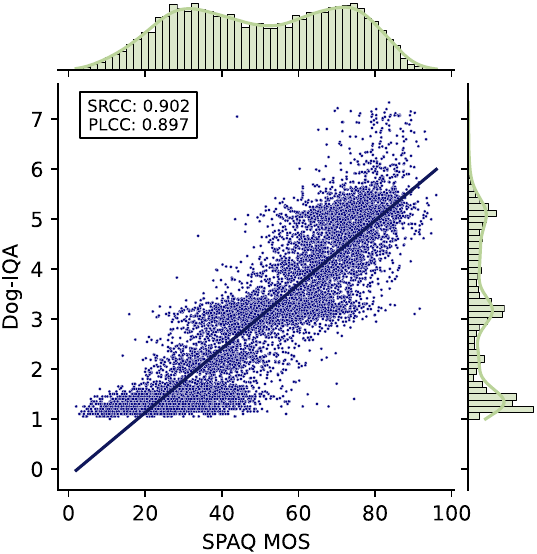}
    \end{minipage}
    \hfill
    \begin{minipage}{0.48\textwidth}
        \centering
        \includegraphics[width=\linewidth]{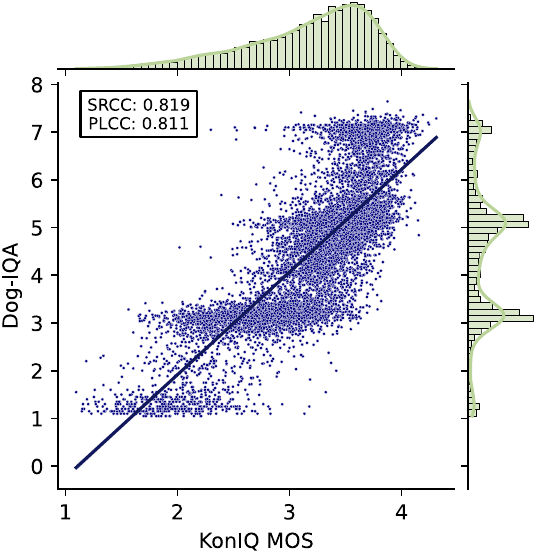}
    \end{minipage}
    \vspace{-3mm}
    \caption{Correlation between MOS and Dog-IQA's scores on SPAQ and KonIQ.
    The marginal hist plots show the distribution of MOS and Dog-IQA's scores.
    And the points $(s^{*},s_{Dog})$ are scattered in the center.
    The regression line shows a linear correlation between Dog-IQA and human scores.
    }
    \label{fig:cor}
    \vspace{-7mm}
\end{figure}

Training-based methods show variability depending on the dataset used for training. 
For example, SRCC and PLCC scores of Q-Aling on KADID-10k drop significantly when switching the training set from KonIQ to SPAQ, despite both being in-the-wild datasets. 
In contrast, Dog-IQA demonstrates stable performance without any training, highlighting its advantage in terms of generalization and cost-efficiency.
Moreover, scoring AI-generated images has become increasingly critical in the current era of AI advancements.
In the KonIQ $\rightarrow$ AGIQA-3k scenario, Dog-IQA achieves the highest SRCC (0.823) and PLCC (0.797), clearly outperforming the second-best model, which only achieves 0.735 SRCC. 
This result underscores the superiority of Dog-IQA in cross-dataset evaluations, especially for AI-generated content.
However, in the SPAQ $\rightarrow$ KonIQ case, Dog-IQA performs slightly lower than Q-Align, which secures the highest SRCC (0.848) and PLCC (0.879). 
Despite this, Dog-IQA still secures the second-best performance, highlighting its robustness.

In conclusion, the analyses of Dog-IQA’s performance across various cross-dataset settings clearly indicate that it achieves SOTA results in most scenarios.
While Q-Align performs slightly better on specific datasets, Dog-IQA’s ability to consistently rank at the top or near the top positions across all datasets demonstrates its robustness and effectiveness as a training-free IQA model.
\vspace{-2mm}
\subsection{Visualizatoin}
\vspace{-2mm}
We visualize the scores predicted by humans and our proposed Dog-IQA on SPAQ and KonIQ datasets in Figure~\ref{fig:cor}.
The range of the final score varies between 1 to 7.66 (SPAQ) and 7.64(KonIQ) which are slightly higher than 7.
This is because the final score consists of the area-weighted average of scores and the number of masks.
As the scores from MLLM are discrete, the final scores are denser around the integer values.
The mask number scheme and area average mechanisms help the continuous-like distribution, which further improve Dog-IQA's performance.

\begin{figure}[t]
    \centering
    \includegraphics[width=\linewidth]{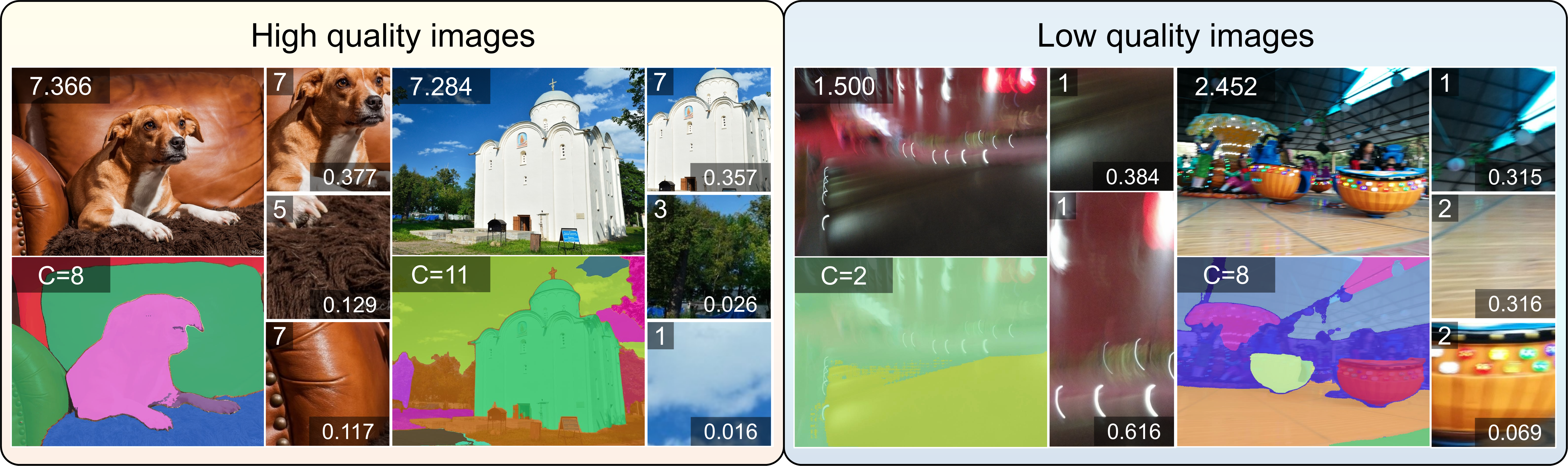}
    \vspace{-6mm}
    \caption{Example images with their segmented images.
    We select images with various scores to present the \textbf{Dog-IQA}'s ability.
    The upper left number is the score while the lower right number is the area.
    The number of masks is shown in the upper left part in the segmented image.
    }
    \label{fig:vis-images}
    \vspace{-6.5mm}
\end{figure}

Figure~\ref{fig:vis-images} shows example segmentations and scoring results.
For the high-quality and low-quality images scored by our Dog-IQA model, we have selected two of each for display.
For each image, the following figures are provided: the full image, the segmented image, and three exemplary masks.
The upper left corner of the full image displays the final score predicted by our Dog-IQA model.
Directly below the full image, the segmentation results are shown, with the mask count indicated in the upper left.
On the right, three masks of varying quality are presented. Each mask is annotated with their corresponding scores (upper left) and area weights (lower right).
From these example figures, we can directly perceive the model's segmentation performance and come to the following conclusions.
% Additionally, the following conclusions can be obtained.

\textbf{First}, the high-quality images tend to be segmented into more sub-images compared with images with motion blur or out-of-focus. 
This indicates that our design of $s_{seg}$ is effective, leading to more pronounced score differences between high and low quality.
\textbf{Second}, by incorporating image segmentation, MLLM is capable of capturing local distortions within the images.
This allows assigning scores to different regions that correspond to their quality, rather than relying on a single overall score. 
This enables MLLM to achieve more precise and human-aligned quality perception.
In conclusion, Dog-IQA could provide accurate and robust scores for different quality levels of images.
\vspace{-2mm}
\subsection{Ablation Study}
\vspace{-2mm}
The ablation studies provided in Tables~\ref{tab:ablation-1},~\ref{tab:ablation-2}, and ~\ref{tab:ablation-3} highlight the significance of various components in our proposed Dog-IQA model. 
By systematically altering key aspects of the model, the experiment evaluates how each component affects performance on two datasets: SPAQ and AGIQA-3k.
We examine components including  1) the standard given to MLLM, 2) the selection of the mask and bounding box, 3) the aggregation of local scores, 4) the effectiveness of $s_{seg}$, 5) the influence of global and local quality, 6) the number of words, and 7) MLLM selection.
The experiment results are shown in Tables~\ref{tab:ablation-1},~\ref{tab:ablation-2}, and ~\ref{tab:ablation-3}.
Next, we will analyze the impact of each component in detail.

\textbf{Standard.}
Standard-guided scoring is a critical aspect of our model.
We compare three forms of standards, namely number, word, and sentence.
The number-based approach involves asking the MLLM to score image quality directly in the range $\{1,2,\dots,K\}$.
The word-based approach adds descriptive adjectives, such as \textit{excellent}, \textit{fair}, and \textit{bad}, to each score.
The sentence-based approach assigns a sentence describing quality for each score level, such as \textit{4: Fair! The overall quality of the image is fair. There are certain merits but also some deficiencies.}

As shown in experiments 1, 2, and 7 in Table~\ref{tab:ablation-1}, the word-based standard yields the best performance as it provides an accurate mapping between number and quality.
While sentences offer more detailed context than numbers, they can introduce abstract terms (\eg \textit{some}, \textit{certain}) that may distract the model, resulting in slightly lower performance.
Numbers, on the other hand, perform poorly because the MLLM struggles to understand their relationship to image quality without additional context.
In conclusion, associating a word with each score effectively enhances the MLLM’s scoring accuracy.

\textbf{Mask and Bounding Box.} 
When scoring sub-images, we test three input formats: masks (semantic object coverings), bounding boxes (enclosing the masks), and the entire image.
As shown in experiments 4, 5, and 7 in Table~\ref{tab:ablation-1}, using masks significantly degrades performance. 
This is mainly because the constant padding applied to masked areas is still interpreted by the MLLM's visual encoder, negatively influencing the score. 
Conversely, using the entire image as input provides moderate results, though still inferior to bounding boxes. 
Bounding boxes improve performance without computational overhead as the padding is always calculated by the visual encoder. 
Therefore, applying bounding boxes as segmentation method is necessary for maximizing Dog-IQA's accuracy.

\begin{table}[t]
\centering
\small
\renewcommand\arraystretch{1.2}
\setlength{\tabcolsep}{1.5mm}
\caption{Ablation study of our proposed Dog-IQA on SPAQ and AGIQA-3k. We test the influence of the aggregation method, segmentation method, standard given to MLLM and the addition of $s_{seg}$. By comparison, our key designs are significant in improving MLLM scoring accuracy.}
\vspace{-3mm}
% \resizebox{\linewidth}{!}{
\begin{tabular}{c:c:c:c:c:cccc}
\toprule
\rowcolor{color3} \multicolumn{5}{c}{Settings} & \multicolumn{2}{c}{SPAQ}  &\multicolumn{2}{c}{AGIQA-3k}   \\ 
\hline
\rowcolor{color3} Exp index & Aggregation  &Segmentation&Standard&$s_{seg}$&SRCC $\uparrow$&PLCC $\uparrow$ & SRCC $\uparrow$&PLCC $\uparrow$   \\
\hline

1 & Area & BBox & Number & $\checkmark$     & 0.764 & 0.756 & 0.633 & 0.618  \\
2 & Area & BBox & Sentence & $\checkmark$   & 0.836 & 0.829 & 0.662 & 0.652  \\
3 & Mean & BBox & Word & $\checkmark$       & 0.767 & 0.740 & 0.781 & 0.683  \\
4 & Area & Mask & Word & $\checkmark$       & 0.715 & 0.669 & 0.684 & 0.615  \\
5 & N/A   & Whole & Word & N/A          & 0.858 & 0.855 & 0.764 & 0.760  \\
6 & Area & BBox & Word & $\times$           & 0.884 & 0.861 & 0.799 & 0.779  \\
7 & Area & BBox & Word & $\checkmark$       & 0.885 & 0.875 & 0.809 & 0.797  \\
8 & Area & BBox+Whole & Word & $\checkmark$       & \textbf{0.902} & \textbf{0.897} & \textbf{0.823} & \textbf{0.797}   \\
\bottomrule
\end{tabular}
% }
\label{tab:ablation-1}
\vspace{-6mm}
\end{table}

\textbf{Score Aggregation.}
We evaluate two score aggregation methods: simple average and area-weighted average.
Considering that the summation of the area should be the area of the image, we use the mask area instead of the sub-image area.
As experiments 3 and 7 in Table~\ref{tab:ablation-1} indicate, there is a significant improvement in both datasets with area-weighted average.
This can be explained by the attention scheme.
There are plenty of small objects that are often scored with low quality because of a lack of pixels.
However, the quality of the image is always represented by the main object, which usually has a larger area.
So more attention should be put on larger objects, namely taking the area-weighted average on quality scores of sub-images, which is more consistent with humans.
In conclusion, leveraging the area-weighted average effectively improves Dog-IQA's accuracy.

\textbf{Effectiveness of $s_{seg}$.}
From experiments 6 and 7 in Table~\ref{tab:ablation-1}, the existence of $s_{seg}$ could assist in image quality assessment.
After adding $s_{seg}$, the PLCC of SPAQ and AGIQA-3k increases by 0.014 and 0.019 respectively.
Because the image quality could influence the performance of most vision models, the performance of the segmentation model could represent image quality.
Therefore, $s_{seg}$ to some extent represents the performance of the segmentation model and accounts for quality assessment.
Although the segmentation model is totally task irrelevant, it still can provide a rough IQA score.
We further test that with only $s_{seg}$, the SRCC on SPAQ could reach around 0.2.
In conclusion, $s_{seg}$ has no computational overhead but still can improve Dog-IQA's performance.

\textbf{Global and Local Quality.}
To validate the significance of local quality versus global quality, we conduct experiments 5, 7, and 8, with results presented in Table~\ref{tab:ablation-1}. 
From the experimental results, we can draw two critical conclusions. 
\textbf{First}, the sum of the quality information from various local sources exceeds the overall information.
Local quality gains higher SRCC (0.885) on SPAQ than global quality (0.858).
This observation highlights the effectiveness of our fine-grained evaluation methodology and the innovative design of our score aggregation process.
\textbf{Second}, although neither the local scores nor the overall score reaches 0.9, averaging the two can still further enhance the model's accuracy. 
For simplicity, we take the mean value of global and local scores.
In summary, the experimental results strongly support the notion that the integration of both global and local quality, namely mix-grained, yields superior results compared to the isolated performance of each.

\begin{wraptable}{r}{0.5\textwidth}
\vspace{-4mm}
\centering
\small
\renewcommand\arraystretch{1.2}
\caption{Number of words (K).}
\vspace{-3mm}
\setlength{\tabcolsep}{1.1mm}
\resizebox{\linewidth}{!}{
\begin{tabular}{l|cccccc}
\toprule
\rowcolor{color3}  & \multicolumn{2}{c}{SPAQ} &\multicolumn{2}{c}{KADID-10k} &\multicolumn{2}{c}{AGIQA-3k}  \\ \cline{2-7}
\rowcolor{color3} \multirow{-2}{*}{K}& SRCC $\uparrow$&PLCC $\uparrow$ & SRCC $\uparrow$&PLCC $\uparrow$ & SRCC $\uparrow$&PLCC $\uparrow$ \\
\hline
3 & 0.731               & 0.722                 & 0.447                 & 0.473             & 0.747             & 0.757 \\
5 & 0.853               & 0.860                 & 0.572                 & 0.576             & \textbf{0.808}    & \textbf{0.797} \\
7 & \textbf{0.885}      & \textbf{0.875}        & \underline{0.580}     & \textbf{0.589}    & \underline{0.800} & \underline{0.779} \\
9 & \underline{0.875}   & \underline{0.840}     & \textbf{0.583}        & \underline{0.586} & 0.743             & 0.753\\

\bottomrule
\end{tabular}
}
\label{tab:ablation-2}
\vspace{-4mm}
\end{wraptable}

\textbf{Number of Words.}
As discussed before, after applying a discrete scoring form, the number of levels decides the performance upper bound of IQA models.
So we test the performance of our proposed Dog-IQA with 3, 5, 7, and 9 words.
All numbers are odd because there needs to be a level representing medium to conform to human evaluation.
The result is shown in Table~\ref{tab:ablation-2}.
Only three words are not enough to gain excellent performance while it still surpasses most of the previous training-free methods (see Table~\ref{tab:comp-no-training}).
Interestingly, the results also indicate that increasing the number of levels beyond a certain point does not necessarily lead to better performance.
Specifically, using 7 words yields the best results in most scenarios and the second-best in the remaining cases.
In summary, 7 appears to be the optimal number of word levels to accurately assess image quality.

\begin{wraptable}{r}{0.45\textwidth}
\centering
\footnotesize
\renewcommand\arraystretch{1.2}
\caption{MLLMs' performance on SPAQ.}
\vspace{-3mm}
\resizebox{\linewidth}{!}{\begin{tabular}{l|cc}
\toprule
\rowcolor{color3} MLLM & SRCC $\uparrow$ & PLCC $\uparrow$  \\
\hline
InternLM-XComposer-1.0 & 0.054 & 0.056 \\
InternLM-XComposer-2.0 & 0.383 & 0.347 \\
LLaVA-v1.5-7b	 & 0.006 & 0.001 \\
LLaVA-v1.5-13b	 & 0.234 & 0.235 \\
LLaVA-Next	 & 0.450 & 0.454 \\
mPLUG-Owl	 & 0.389 & 0.386 \\
mPLUG-Owl2	 & 0.347 & 0.346 \\
mPLUG-Owl3	 & \textbf{0.858} & \textbf{0.855} \\
\bottomrule

\end{tabular}
}
\label{tab:ablation-3}
\vspace{-3mm}
\end{wraptable}

\textbf{MLLM Selection.}
Given the critical role of MLLMs in scoring, we evaluate some MLLMs' performance on SPAQ.
The MLLMs includes InternLM-XComposer-1.0~\citep{zhang2023internlm}, InternLM-XComposer-2.0~\citep{dong2024internlm}, LLaVA-v1.5-7b~\citep{liu2024visual}, LLaVA-v1.5-13b~\citep{liu2024visual}, LLaVA-Next~\citep{liu2024llava}, mPLUG-Owl~\citep{ye2023mplug}, mPLUG-Owl2~\citep{ye2024mplug_2}, and mPLUG-Owl3~\citep{ye2024mplug_3}.
All models are tested using 5 words with the whole image as input, reflecting their fundamental zero-shot IQA capabilities.
The results are shown in Table~\ref{tab:ablation-3}, from which we can draw the following conclusions.
\textbf{First}, from the version's perspective, the trend shows that the higher the version is, the better the model's performance is.
\textbf{Second}, when we consider different models, mPLUG-Owl3 demonstrates a clear performance advantage, and LLaVA-Next gains sub-optimal performance.
Therefore, we choose mPLUG-Owl3 as our scoring model.
\vspace{-3mm}
\section{Limitations and Discussions}
\vspace{-3mm}
In this section, we will discuss the limitations of our proposed Dog-IQA.
\textbf{First}, the impressive performance of Dog-IQA can be attributed not only to our novel design but also to the capabilities of the underlying MLLM. 
Ultimately, it is the MLLM that generates the quality scores, while our design better exploits its extensive prior knowledge. 
However, as shown in Table~\ref{tab:ablation-3}, the performance of MLLM increases significantly with version updation which will finally promote the development of IQA. 
Consequently, the selection of MLLM matters and Dog-IQA's performance may decline when utilizing MLLMs with poor image understanding ability.

\textbf{Second}, as an important part of the overall pipeline, the segmentation process can significantly impact the accuracy of Dog-IQA. 
If we switch to a segmentation model with subpar performance, mix-grained segmentation may fail, resulting in a direct score for the entire image instead. 
Additionally, if the segmentation model primarily outputs bounding boxes that lack a clear main object—such as only capturing the edges of an object or half of a human face—this can lead to MLLM's misjudgment and inaccurate scores, further degrade our Dog-IQA's performance.
Thus, the choice of segmentation model and the segmentation granularity are critical factors influencing Dog-IQA's performance.

\textbf{Third}, because the MLLM must evaluate the quality of each mask, the inference speed of Dog-IQA is relatively slow compared to models that require only a single inference. 
On average, Dog-IQA processes 7.22 masks and the entire image, resulting in 7$
\times$ longer inference time. % than when evaluating the entire image at once. 
After testing on a single NVIDIA RTX A6000 GPU, our proposed Dog-IQA can segment the whole SPAQ dataset in 50 minutes and score each mask and the total data within 6 hours.
This process can be performed with data parallel, which means it takes around 1.5 hours to obtain the final result when running on 4 GPUs.
% As an MLLM with 8B parameters, the inference speed is relatively fast.
While the text embeddings can be pre-calculated and reused, allowing for the omission of the text encoder, the total inference time remains longer than single forward inference.
In conclusion, Dog-IQA may suffer from low processing speed if the segmentation granularity is too fine.
\vspace{-3mm}
\section{Conclusion} 
\vspace{-3mm}
In this work, we propose Dog-IQA, a standard-guided zero-shot mix-grained IQA method, which is training-free and utilizes the exceptional prior knowledge of MLLMs.
With the combination of SAM2 and mPLUG-Owl3, we propose two key mechanisms to enhance IQA performance. 
The standard-guided scoring mechanism ensures consistent and objective quality evaluation by aligning scores with predefined standards. 
The mix-grained aggregation mechanism refines the final quality score by aggregating global and object-centered sub-image quality scores. 
We conduct extensive experiments across a variety of datasets, benchmarking our proposed Dog-IQA against SOTA methods. 
The results demonstrate that Dog-IQA outperforms all previous training-free approaches and achieves competitive performance relative to training-based methods, which strongly supports the novelty and robustness of our proposed mechanisms.
We also systematically conduct ablation studies, which further confirm the effectiveness of the novel mechanisms.
This work highlights the exceptional image understanding capabilities of MLLMs and confirms the feasibility of attaining remarkable outcomes using solely pre-trained models. 
Future research will aim to reduce the computational costs associated with multiple inferences and enhance pixel-level quality assessments.

\newpage

% \bibliography{iclr2025_conference}
\bibliographystyle{iclr2025_conference}

\end{document}